\documentclass[journal]{IEEEtran}
\usepackage{latexsym,,amssymb,amsmath,graphicx,epsf,cite,bbm,float}
\usepackage{ifpdf}
\usepackage{epstopdf,mathtools}

\usepackage{algorithm,algorithmic}
\usepackage{amsmath,amssymb,bm}
\usepackage{amsfonts,dsfont,color,bbm,subcaption}

\newcommand{\abs}[1]{|#1|}
\newcommand{\bs}[1]{\boldsymbol{#1}}

\DeclareMathOperator*{\argmin}{arg\,min}

\usepackage{hyperref,amsthm}

\newtheorem{theorem}{Theorem}

\newtheorem{lemma}[]{Lemma}

\newtheorem{proposition}[]{Proposition}

\newtheorem{remark}[]{Remark}

\newtheorem{definition}[]{Definition}

\begin{document}

\title{Formulation of Weighted Average Smoothing as a Projection of the Origin onto a Convex Polytope} 
\author{\IEEEauthorblockN{Kaan Gokcesu}, \IEEEauthorblockN{Hakan Gokcesu} }
\maketitle

\begin{abstract}
	Our study focuses on determining the best weight windows for a weighted moving average smoother under squared loss. We show that there exists an optimal weight window that is symmetrical around its center. We study the class of tapered weight windows, which decrease in weight as they move away from the center. We formulate the corresponding least squares problem as a quadratic program and finally as a projection of the origin onto a convex polytope. Additionally, we provide some analytical solutions to the best window when some conditions are met on the input data.
\end{abstract}

\section{Introduction}

Smoothing is a useful technique in machine learning to identify patterns in a dataset and eliminate noise or outliers \cite{hardle1991smoothing}. 
The underlying assumption is that nearby data points, in some order (such as time for temporal data), are more closely related than those farther apart. Smoothing methods are widely available in the literature and can produce more robust and flexible analyses of datasets by generating informative, smoothed versions of the vanilla input \cite{simonoff2012smoothing}. 
Smoothing has diverse applications in many fields such as data analysis \cite{kenny1998data,brandt1998data,guo2004functional,deveaux1999applied,zin2020smoothing}, signal processing \cite{roberts1987digital,orfanidis1995introduction,fong2002monte,schaub2018flow}, anomaly detection \cite{chandola2009anomaly,gokcesu2017online,tsopelakos2019sequential, gokcesu2018sequential}, and machine learning \cite{saul2004overview,gokcesu2020recursive,yaroshinsky2001smooth,gokcesu2020generalized,servedio2003smooth,gokcesu2021optimal,hartikainen2010kalman, gokcesu2021optimally}, where it can provide more informative data and performance gains.

Smoothing and curve-fitting are two related but distinct fields \cite{arlinghaus1994practical}. While curve-fitting aims to determine a functional form that best fits the data, smoothing is only concerned with obtaining the smoothed values of the dataset. Smoothing typically involves selecting a parameter that determines the degree of smoothing, whereas curve-fitting involves optimizing inherent parameters for the best fit. Linear smoothers are a widely applied type of smoother that expresses smoothed data points as a weighted average of its nearby data points \cite{buja1989linear}. This approach can be achieved through the convolution of the original signal with a finite impulse response (FIR) filter \cite{oppenheim1997signals}. One popular linear smoother is the moving average, particularly in time-series analysis, which attenuates short-term deviations and emphasizes long-term behavior \cite{chatfield2013analysis,wei2006time,hansun2013new}. The decision of what constitutes short or long-term behavior determines the smoother parameters, such as the length of the moving average window \cite{hatchett2010optimal}. Moving average is a commonly used tool in mathematical finance to analyze stock prices and trading volumes, as well as in macroeconomic metrics such as GDP, imports, exports, and unemployment \cite{chiarella2006dynamic,schwert1987effects,gokcesu2021nonparametric}. It also has applications in ECG analysis, particularly in QRS detection 
\cite{chen2003moving}. Moving average can be achieved through the convolution of the original signal with a probability distribution, effectively serving as a low-pass filter to filter out high-frequency components and smooth the signal \cite{kaiser1977data}.

The moving median is another type of the smoothing techniques, which differs from the moving average  \cite{justusson1981median}. While the moving average is statistically optimal for removing Gaussian noise from observations \cite{kalman1960new}, it is not effective when the noise is non-Gaussian, such as when it has a heavy tail or contains outliers. This is because the moving average is sensitive to such outliers, which can negatively affect the accuracy of the smoothed data. To address this issue, more robust techniques are required \cite{gokcesu2022nonconvex,gokcesu2021generalized,khargonekar1985non,gokcesu2021regret}, such as the moving median. The moving median is a nonlinear smoother, which calculates the smoothed values by taking the median of nearby samples. This technique is statistically optimal when the noise in the input signal follows a Laplace distribution \cite{arce2005nonlinear}. One advantage of the moving median is that it is more effective at preserving the edges of an image compared to other smoothing techniques, which makes it useful in image processing applications \cite{ataman1981some}. It is also used in edge detection \cite{huang1998computing} and mass spectrometry \cite{do1995applying}. The skiplist data structure can be used to efficiently compute the moving median, making it computationally efficient \cite{pugh1990skip}.

In summary, smoothing techniques deal with creating a smoothed signal $\{x_n\}_{n=1}^N$ from an observation sequence $\{y_n\}_{n=1}^N$ using a sliding window \cite{lee2001sliding} of some length $2K+1$. The smoothed signal is computed as a function of the values within the sliding window, i.e., 
\begin{align}
	x_n=\mathcal{T}(\{y_m\}_{m=n-K}^{n+K}).
\end{align}
This function $\mathcal{T}(\cdot)$ can be a mean, median, or some other function, and can also be weighted if a window function is used, i.e.,
\begin{align}
	x_n=\mathcal{T}(\{y_m,w_{m-n}\}_{m=n-K}^{n+K}).
\end{align} 
The window weights $\{w_k\}_{k=-K}^{K}$ determine the weighting of the values within the sliding window and is an important aspect of the smoothing technique.

Window functions \cite{weisstein2002crc} are used in various fields such as spectral analysis \cite{stoica2005spectral}, antenna design \cite{rudge1982handbook}, and beam-forming \cite{van1988beamforming}. However, in statistical analysis \cite{dixon1951introduction}, window functions are used to define a weighting vector for the smoothing of a dataset. In curve fitting \cite{o1978curve} and Bayesian analysis \cite{ghosh2006introduction}, they are also referred to as kernels \cite{keerthi2003asymptotic}. Kernel smoothing \cite{wand1994kernel} is a statistical technique that creates data estimates as a weighted average of neighboring observations, where weights are defined by a kernel. The weighting function of a kernel is designed such that it decreases from its peak in all directions, meaning that closer data points are given higher weights. In this regard, we focus on the problem of optimal weight design and show that it can be formulated as a projection problem.


\section{Preliminaries}\label{sec:problem}
\subsection{Problem Definition}
We start with some preliminaries. Let us have the following observed input signal samples 
\begin{align}
	\boldsymbol{y}=\{y_n\}_{n=1}^{N}.
\end{align}
We smooth the input vector $\boldsymbol{y}$ and construct our smoothed vector
\begin{align}
	\boldsymbol{x}=\{x_n\}_{n=1}^{N}.
\end{align}
We start by focusing on the design of a moving average smoother. Hence, to create $\boldsymbol{x}$, we pass $\boldsymbol{y}$ through a weighted moving average (weighted mean) filter $\boldsymbol{w}=\{w_k\}_{k=-M}^{M}$ of window size $2M+1\leq N$, where $\boldsymbol{w}$ is in a probability simplex. Hence, the weights are positive and sum to $1$, i.e.,
\begin{align}
	w_k\geq& \enspace0, &&k\in\{-M,\ldots,M\},\\
	\sum_{k=-M}^{M}w_k=&\enspace1.
\end{align}
Let $N$ be odd, i.e., $N=2K+1$ for some natural number $K$. Then, without loss of generality, we can define the weight vector $\boldsymbol{w}$ for a window length of $N=2K+1$ with some trailing zeros, i.e.,
\begin{align}
	\tilde{w}_k=\begin{cases}
		w_k,&-M\leq k\leq M\\
		0,& \text{otherwise}
	\end{cases},
\end{align} 
for $k\in\{-K,\ldots,K\}$. Since $\sum_{k=-K}^K\tilde{w}_k=1,$ it is still a probability distribution. Thus, without loss of generality, we can assume the window length is $N$.
The elements of the smoothed signal $\boldsymbol{x}$ is calculated from the input $\boldsymbol{y}$ via
\begin{align}
	x_n=\sum_{k=-K}^{K}w_ky_{n+k},\label{eq:wmean}
\end{align}  
where the weighted mean filter is cyclic, i.e., 
\begin{align}
	y_{n+k}=y_{(n+k-1)\pmod N+1}.\label{eq:ycyclic}
\end{align}
Before investigating the smoothness of $\boldsymbol{{x}}$, we first need to define its similarity to the original input signal $\boldsymbol{{y}}$.
We define the similarity performance of this weighted moving average filter by its cumulative square error, i.e.,
\begin{align}
	L(\boldsymbol{w})=&\sum_{n=1}^{N}\left(y_{n}-x_n\right)^2,
	\\=&\sum_{n=1}^{N}\left(y_{n}-\sum_{k=-K}^{K}w_ky_{n+k}\right)^2.\label{eq:Lw}
\end{align}
However, a question arises about which $\boldsymbol{w}$ is most suitable to smooth the given input observation vector $\boldsymbol{y}$. To find the optimal $\boldsymbol{w}$, we minimize the objective function in \eqref{eq:Lw} with respect to $\boldsymbol{w}$, i.e.,
\begin{align}
	\argmin_{\boldsymbol{w}\in\mathcal{W}}L(\boldsymbol{w})
\end{align} 
where $\mathcal{W}$ is a subset of the probability simplex of dimension $N$, i.e., $\mathcal{P}^N$.

\subsection{Window Design and Non-trivial Solutions}
For the given objective in \eqref{eq:Lw}, we have the following trivial solution.
\begin{remark}
	Note that the cumulative loss function $L(\boldsymbol{w})$ is nonnegative since it is a sum of squares. When the set $\mathcal{W}$ is the whole probability simplex $\mathcal{P}^N$, it has the following trivial minimizer $\boldsymbol{w^0}=\{w_k^0\}_{k=-K}^K$:
	\begin{align}
		w_k^0=\begin{cases}
			1,& k=0\\
			0,& k\neq0 
		\end{cases},
	\end{align}
	since $L(\boldsymbol{w^0})=0$.
\end{remark}
In fact, we have the following result that show the dominant effect the center weight $w_0$ has.
\begin{proposition}
	For any $\boldsymbol{\tilde{w}}=\lambda\boldsymbol{w}+(1-\lambda)\boldsymbol{w^0}$, we have
	\begin{align*}
		L(\boldsymbol{\tilde{w}})\leq L(\boldsymbol{w}),
	\end{align*}
	where $0\leq \lambda\leq 1$.
\begin{proof}
	For any $\boldsymbol{w}$, we have
		\begin{align}
		L(\boldsymbol{\tilde{w}})=&L(\lambda\boldsymbol{w}+(1-\lambda)\boldsymbol{w^0}),\\
		=&\sum_{n=1}^{N}\left(y_{n}-\sum_{k=-K}^{K}(\lambda w_k+(1-\lambda)w^0_k)y_{n+k}\right)^2,\\
		=&\sum_{n=1}^{N}\left(\lambda y_{n}-\sum_{k=-K}^{K}\lambda w_ky_{n+k}\right)^2,\\
		=&\lambda^2\sum_{n=1}^{N}\left( y_{n}-\sum_{k=-K}^{K} w_ky_{n+k}\right)^2,\\
		=&\lambda^2 L(\boldsymbol{w}),
	\end{align}
	which concludes the proof.
\end{proof}
\end{proposition}
Hence, increasing the center weight always decreases the cumulative loss. To effectively eliminate this triviality, we incorporate only the neighboring samples in our smoothing by setting the center weight to $0$, i.e., $$w_0=0.$$ 

Moreover, following the traditional literature, we constrain our weight vectors to have tapering weights, i.e., it should be nonincreasing away from the center. This is intuitive since the elements of the resulting smoothed signal should be more influenced by the closer neighbors. Thus, for every $\boldsymbol{w}\in\mathcal{W}$, we have
\begin{align}
	\boldsymbol{{w}}:
	\begin{cases}
		w_0=0&\\
		w_j\geq w_k,& 0< j\leq k\\
		w_j\leq w_k,& j\leq k < 0
	\end{cases},\label{eq:taper}
\end{align}
where the window $\boldsymbol{{w}}$ diminishes with the distance to the center.

We reformulate our objective in \eqref{eq:Lw} over a unit simplex in the next section.
\section{Problem Reformulation}
\subsection{Convex Programming}
To compactly reformulate the problem definition in \eqref{eq:Lw}, we first define the autocorrelation values of the input signal $\boldsymbol{{y}}.$
\begin{definition}\label{def:r}
	The autocorrelation of $\boldsymbol{y}$ is defined as follows:
	\begin{align*}
		r_{t}=\sum_{n=1}^{N}y_ny_{n+t},
	\end{align*} 
	where ${y}_n$ is cyclic as in \eqref{eq:ycyclic}.
\end{definition}
Given \autoref{def:r}, we have the following properties.
\begin{proposition}\label{thm:rt}
	From \autoref{def:r} and \eqref{eq:ycyclic}, we have the following properties:
	\begin{itemize}
		\item $r_t$ is symmetric around $0$, i.e., $r_t=r_{-t}$.
		\item $r_t$ is periodic with $N$, i.e., $r_{t}=r_{N+t}$.
		\item For any integer $k$, we have $\sum_{n=1}^{N}y_{n+k}y_{n+k+t}=r_t$.
	\end{itemize}
\end{proposition}

Using the properties in \autoref{thm:rt}, we can reformulate the problem in \eqref{eq:Lw} as the following.
\begin{lemma}\label{thm:quad}
	Using \autoref{thm:rt}, we can reduce the optimization problem to the following quadratic programming:
	\begin{align*}
		\argmin_{\boldsymbol{w}\in\mathcal{W}}\boldsymbol{w}^T\boldsymbol{R}\boldsymbol{w}-2\boldsymbol{w}^T\boldsymbol{r},
	\end{align*}
	where 
	$\boldsymbol{w}$ is the weight vector, 
	$\boldsymbol{R}$ is the autocorrelation matrix (i.e., $\boldsymbol{R}(i,j)=r_{j-i}$) and $\boldsymbol{r}=\{r_k\}_{k=-K}^K$ is the autocorrelation vector.
	\begin{proof}
		Using \autoref{thm:rt} and after some algebra, we can alternatively write the objective function $L(\boldsymbol{w})$ as
		\begin{align}
			L(\boldsymbol{w})=&\sum_{n=1}^{N}\left(y_{n}-\sum_{k=-K}^{K}w_ky_{n+k}\right)^2,\\
			=&r_0-2\sum_{k=-K}^{K}w_kr_k+\sum_{k=-K}^{K}\sum_{m=-K}^{K}w_kw_mr_{m-k},\nonumber\\
			=&\boldsymbol{w}^T\boldsymbol{Rw}-2\boldsymbol{w}^T\boldsymbol{r}+r_0,
		\end{align}
		which concludes the proof since $r_0$ is constant.
	\end{proof}
\end{lemma}

\begin{lemma}\label{thm:Rpsd}
	The problem in \autoref{thm:quad} is convex in $\boldsymbol{w}$; since the autocorrelation matrix $\bs{R}$ is positive semi-definite, i.e., for any $\boldsymbol{v}\in\Re^N$, we have
	\begin{align*}
		\boldsymbol{v}^T\boldsymbol{R}\boldsymbol{v}\geq 0.
	\end{align*}
	\begin{proof}
		From \autoref{def:r}, we can write the matrix $\boldsymbol{{R}}$ as a sum of outer products
		\begin{align}
			\boldsymbol{R}=\sum_{n=1}^{N}\boldsymbol{z}_n\boldsymbol{z}_n^T,
		\end{align}
		where the vector $\boldsymbol{z}_n$ is equal to the vector $\boldsymbol{y}$ that is shifted by $n$ in a cyclic manner. Hence,
		\begin{align}
			\boldsymbol{v}^T\boldsymbol{R}\boldsymbol{v}=&\sum_{n=1}^{N}\boldsymbol{v}^T\boldsymbol{z}_n\boldsymbol{z}_n^T\boldsymbol{v}	=\sum_{n=1}^{N}\alpha_n^2,
		\end{align}
		where $\alpha_n$ is the inner product of $\boldsymbol{v}$ with $\boldsymbol{z}_n$. Since sum of squares are nonnegative, $\bs{R}$ is positive semi definite.
	\end{proof}
\end{lemma}

\subsection{Optimization over a Unit Simplex}
\begin{proposition}\label{thm:Lsymm} $\boldsymbol{\tilde{w}}=\{\tilde{w}\}_{k=-K}^K$ be symmetric to $\boldsymbol{w}=\{w_k\}_{k=-K}^K$ around $k=0$, i.e.,
	\begin{align*}
		\tilde{w}_k=w_{-k},&&k\in\{-K,\ldots,K\}.
	\end{align*}
	Then, we have
	\begin{align*}
		L(\boldsymbol{w})=L(\boldsymbol{\tilde{w}}).
	\end{align*}
	\begin{proof}
		Since the autocorrelation vector $\boldsymbol{r}$ is symmetric, we have
		\begin{align}
			\boldsymbol{\tilde{w}}^T\boldsymbol{r}=\boldsymbol{w}^T\boldsymbol{r}.
		\end{align}
		Moreover, since the autocorrelation matrix $\bs{R}$ is symmetric around both its diagonals, we also have
		\begin{align}
			\boldsymbol{\tilde{w}}^T\boldsymbol{R\tilde{w}}=\boldsymbol{w}^T\boldsymbol{Rw},
		\end{align}
		which concludes the proof.
	\end{proof}
\end{proposition}

\begin{proposition}\label{thm:w*symm}
	There exists a minimizer $\boldsymbol{w}^*=\{w^*_k\}_{k=-K}^K\in\mathcal{W}$ for the problem in \autoref{thm:quad} that is symmetric, i.e.,
	\begin{align*}
		w^*_k=w^*_{-k},&& k\in\{-K,\ldots,K\}.
	\end{align*}
	\begin{proof}
		Let $\boldsymbol{w_0}\in\mathcal{W}$ be a minimizer for $L(\boldsymbol{w})$, i.e.,
		\begin{align}
			L(\boldsymbol{w_0})=\min_{\boldsymbol{w}\in\mathcal{W}}L(\boldsymbol{w}).
		\end{align}
		Let $\boldsymbol{\tilde{w}_0}$ be its symmetric weights. From \autoref{thm:Rpsd}, we have
		\begin{align}
			L\left(\frac{1}{2}\boldsymbol{w_0}+\frac{1}{2}\boldsymbol{\tilde{w}_0}\right)\leq&\frac{1}{2}L(\boldsymbol{w_0})+\frac{1}{2}L(\boldsymbol{\tilde{w}_0}),\\
			\leq&L(\boldsymbol{w_0}),
		\end{align}
		since $\boldsymbol{\tilde{w}}$ has the same loss as $\boldsymbol{w}$ from \autoref{thm:Lsymm}. Thus, the average of $\boldsymbol{w}$ and its symmetric $\boldsymbol{\tilde{w}}$, which is a symmetric weighting, is also a minimizer and concludes the proof.
	\end{proof}
\end{proposition}

Henceforth, we additionally restrict our class to the set of symmetric weights, i.e., for every $\boldsymbol{w}\in\mathcal{W}$, we have
\begin{align}
	\boldsymbol{{w}}:
	\begin{cases}
		w_0=0&\\
		w_k=w_{-k}& \forall k\\
		w_j\geq w_k,& \forall j,k:0< j\leq k\\
		w_j\leq w_k,& \forall j,k: j\leq k < 0
	\end{cases}.
\end{align}

\begin{remark}\label{def:poly}
	We observe that the class $\mathcal{W}$ is a convex polytope defined by its vertices $\{\boldsymbol{v_i}\}_{i=1}^K$, where $\boldsymbol{v_i}=\{v_{i,k}\}_{k=-K}^K$ is given by
	\begin{align*}
		v_{i,k}=\begin{cases}
			0, &k=0\\
			\frac{1}{2i}, &1\leq\abs{k}\leq i,\\
			0,& \text{otherwise}
		\end{cases}.
	\end{align*}
\end{remark}

When $\mathcal{W}$ is a convex polytope as defined in \autoref{def:poly}, we have the following property.

\begin{proposition}\label{thm:poly}
	From \autoref{def:poly}, we can write any point $\boldsymbol{w}\in\mathcal{W}$ as a convex combination of the vertices $\{\boldsymbol{v_i}\}_{i=1}^K$, i.e.,
	\begin{align*}
		\boldsymbol{w}=\sum_{i=1}^{K}p_i\boldsymbol{v_i}=\boldsymbol{V}\boldsymbol{p},
	\end{align*}
	where $\boldsymbol{p}=\{p_i\}_{i=1}^K$ is in the probability simplex $\mathcal{P}^K$, i.e., 
	$p_i\geq 0$ for $ i\in\{1,\ldots,K\}$ and $\sum_{i=1}^{K}p_i=1$; and the $i^{th}$ column of the matrix $\boldsymbol{V}$ is $\boldsymbol{v_i}$.
\end{proposition}

\subsection{Quadratic Programming}
Next, we reformulate the problem as a quadratic program with respect to $\bs{p}$.
\begin{lemma}\label{thm:QP}
	The problem in \autoref{thm:quad} is equivalent to
	\begin{align*}
		&\text{minimize} &&\frac{1}{2}\boldsymbol{p}^T\boldsymbol{Q}\boldsymbol{p}-\boldsymbol{{b}}^T\boldsymbol{{p}}\\
		&\text{subject to} &&\boldsymbol{{p}}\succeq\boldsymbol{{0}}
		\\& &&\boldsymbol{{e}}^T\boldsymbol{{p}}=1 
	\end{align*}
	where $\boldsymbol{{Q}}=\boldsymbol{{V}}^T\boldsymbol{{RV}}$, $\boldsymbol{{b}}=\boldsymbol{{V}}^T\boldsymbol{{r}}$ and $\bs{e}$ is an all-one vector.
	\begin{proof}
		Since the matrix $\boldsymbol{{V}}$ is predetermined irrespective of $\boldsymbol{{p}}$, we can reformulate the problem in terms of the distribution $\boldsymbol{{p}}.$
		The new objective is given by 
		\begin{align}
			L_Q(\boldsymbol{p})=\boldsymbol{p}^T\boldsymbol{V}^T\boldsymbol{RVp}-2\boldsymbol{p}^T\boldsymbol{V}^T\boldsymbol{r},
		\end{align}
		which is convex in $\boldsymbol{p}$ and is a quadratic program over the unit simplex. 
	\end{proof}
\end{lemma}

This problem can be solved using a variety of techniques, including gradient descent, interior point methods, where the optimal distribution $\boldsymbol{{p_*}}$ can be calculated using various iterative methods approximately with varying degrees of convergence \cite{boyd2004convex}. However, for an analytical solution, we need to investigate the properties of the specific problem at hand. 

\begin{proposition}
	The unconstrained optimizer of the objective in \autoref{thm:QP} is given by
	$$\boldsymbol{{p}}_{Q}=\boldsymbol{{(V^TRV)}^{-1}V^Tr}.$$
	\begin{proof}
		By setting the gradient to zero, we have
		\begin{align}
			\nabla L_Q(\bs{p})=&\bs{Qp}_Q-\bs{b}=0,
			\\\boldsymbol{{p}}_{Q}=&\boldsymbol{{Q}^{-1}b},
		\end{align}
	which concludes the proof.
	\end{proof}
\end{proposition}
If $\boldsymbol{{p}}_Q$ is in the unit simplex, we have reached a minimizer. Otherwise, we can utilize the method of Lagrange multipliers to enforce the equality constraint with an additive loss term $\lambda(1-\boldsymbol{e^Tp})$.
\begin{proposition}
	The optimizer of the objective in \autoref{thm:QP} when $\bs{e^Tp=1}$ is given by
	\begin{align*}
		\bs{p}_{\lambda_*}=\bs{p}_Q+\frac{1-\bs{e^Tp}_Q}{\bs{e^T(V^TRV)^{-1}e}}\bs{(V^TRV)^{-1}e}.
	\end{align*}
	\begin{proof}
		By setting the gradient of the Lagrangian function to zero, we have
		$$\boldsymbol{p}_{\lambda}=\boldsymbol{Q^{-1}}(\bs{b}+\lambda \bs{e}).$$ From the equality constraint, we have $\bs{e^Tp_{\lambda}}=1$. Hence, 
		\begin{align}
			\lambda_*=&\frac{1-\bs{e^TQ^{-1}b}}{\bs{e^TQ^{-1}e}},
			\\\bs{p}_{\lambda_*}=&\bs{Q^{-1}b}+\frac{1-\bs{e^TQ^{-1}b}}{\bs{e^TQ^{-1}e}}\bs{Q^{-1}e},
		\end{align}
		which concludes the proof.
	\end{proof}
\end{proposition}
  
If $\bs{p}_{\lambda_*}\succeq 0$, we have reached a minimizer. Otherwise, we need to enforce the inequalities with an additive term $-\bs{\mu^Tp}$ to the Lagrangian function.
However, solving this Lagrangian function is not straightforward.

\subsection{Minimum Norm Problem}
Next, we reformulate the problem as a minimum Mahalanobis norm problem.
\begin{lemma}\label{thm:MN}
	The problem in \autoref{thm:QP} is equivalent to
	\begin{align*}
		&\text{minimize} &&\bs{p^T\tilde{Q}p}\\
		&\text{subject to} &&\boldsymbol{{p}}\succeq\boldsymbol{{0}}
		\\& &&\boldsymbol{{e}}^T\boldsymbol{{p}}=1 
	\end{align*}
	where $\bs{\tilde{Q}}=\bs{\tilde{V}^TR\tilde{V}}$, $\bs{\tilde{V}}=\bs{v_0e^T}-\bs{V}$ and $\bs{v_0}=\{v_{0,k}\}_{k=-K}^K$ such that $v_{0,0}=1$ and $v_{0,k}=0$ for $k\neq 0$.
	where $\boldsymbol{{Q}}=\boldsymbol{{V}}^T\boldsymbol{{RV}}$, $\boldsymbol{{b}}=\boldsymbol{{V}}^T\boldsymbol{{r}}$ and $\bs{e}$ is an all-one vector.
	\begin{proof}
		We observe that the original objective was a least squares one since $\bs{R}=\bs{Y^TY}$, $\bs{r}=\bs{Y^Ty}$, where the columns of the matrix $\bs{Y}$ are the shifted versions of the input signal $\bs{y}$. We have
		\begin{align}
			L(\boldsymbol{p})=\|\bs{y}-\bs{YVp}\|_2^2.
		\end{align}
		By utilizing $\bs{e^Tp}=1$, we can write the objective as
		\begin{align}
			L(\bs{p})=&\|\bs{Y\tilde{V}p}\|_2^2=\bs{p^T\tilde{V}^TR\tilde{V}p},
		\end{align}
		where $\bs{\tilde{V}}=\bs{v_0e^T}-\bs{V}$ and $\bs{v_0}=\{v_{0,k}\}_{k=-K}^K$ such that $v_{0,0}=1$ and $v_{0,k}=0$ for $k\neq 0$.
	\end{proof}
\end{lemma}

\begin{proposition}
	The optimizer of the objective in \autoref{thm:MN} when $\bs{e^Tp=1}$ is given by
	\begin{align*}
		\bs{p}_{M,\lambda_*}=\frac{\bs{(\tilde{V}^TR\tilde{V})^{-1}e}}{\bs{e^T(\tilde{V}^TR\tilde{V})^{-1}e}}.
	\end{align*}
	\begin{proof}
		Let $\bs{\tilde{Q}}=\bs{\tilde{V}^TR\tilde{V}}$. 
		We utilize the Lagrange multipliers and set the gradient to zero. We have
		\begin{align}
			\bs{p}_{M,\lambda_*}=\frac{\bs{\tilde{Q}^{-1}e}}{\bs{e^T\tilde{Q}^{-1}e}},
		\end{align}
		since $\bs{e^Tp_{B,\lambda_*}}=1$, which concludes the proof.
	\end{proof}
\end{proposition}

If $p_{B,\lambda_*}\succeq0$, we have reached a minimizer. Otherwise, we can reduce the problem to the projection of origin onto a convex polytope.

\begin{theorem}\label{thm:proj}
	The problem in \autoref{thm:MN} is equivalent to
	\begin{align*}
		&\text{minimize} &&\|\bs{x}\|_2\\
		&\text{subject to} &&\bs{x}\in\mathcal{A}
	\end{align*}
	where $\mathcal{A}$ is the convex polytope defined by the columns of $\bs{A}$, where $\bs{A}$ is the square-root of $\bs{\tilde{Q}}$, i.e., $\bs{AA=\tilde{Q}}$.
	\begin{proof}
		We observe that
		\begin{align}
			L(\bs{p})=\|\bs{Ap}\|_2^2,
		\end{align}
		for some $\bs{A}=\bs{Q^{\frac{1}{2}}}$, and $\bs{A}$ is symmetric since $\bs{\tilde{Q}}$ is symmetric and positive semidefinite, which concludes the proof.
	\end{proof}
\end{theorem}

Henceforth, the problem can be reduced to finding the minimum norm point in a convex polytope, i.e., projection of the origin onto a convex polytope. There exist many methods for this problem, e.g., Wolfe's method \cite{wolfe1976finding}. Moreover, this problem has many reductions to other ML problems as detailed in \cite{gabidullina2018problem}.

%
%

\bibliographystyle{IEEEtran}
\bibliography{double_bib}

\end{document}